\documentclass{article}
\usepackage{style/spconf,amsmath,graphicx}
\usepackage{xcolor}
\usepackage{booktabs}
\usepackage{multirow}
\usepackage{caption}
\usepackage{subcaption}
\usepackage{enumitem}
\usepackage[normalem]{ulem}
\usepackage{cite}
\usepackage{float}
\usepackage{amssymb}
\usepackage{pifont}
\newcommand{\cmark}{\ding{51}}%
\newcommand{\xmark}{\ding{55}}%

\title{Pronunciation Assessment with Multi-modal Large Language Models}
\name{Kaiqi Fu$^1$, Linkai Peng$^2$, Nan Yang$^1$, Shuran Zhou$^1$}

\address{
  $^1$Zuoyebang AI Research \\
  $^2$University of Connecticut}

\begin{document}
\ninept
\maketitle
\begin{abstract}
Large language models (LLMs), renowned for their powerful conversational abilities, are widely recognized as exceptional tools in the field of education, particularly in the context of automated intelligent instruction systems for language learning. In this paper, we propose a scoring system based on LLMs, motivated by their positive impact on text-related scoring tasks. Specifically, the speech encoder first maps the learner’s speech into contextual features. The adapter layer then transforms these features to align with the text embedding in latent space. The assessment task-specific prefix and prompt text are embedded and concatenated with the features generated by the modality adapter layer, enabling the LLMs to predict accuracy and fluency scores. Our experiments demonstrate that the proposed scoring systems achieve competitive results compared to the baselines on the Speechocean762 datasets. Moreover, we also conducted an ablation study to better understand the contributions of the prompt text and training strategy in the proposed scoring system.


\end{abstract}

\section{Introduction}
\label{sec:intro}
With the development of globalization, mastering multiple foreign languages has become an essential skill in life and work. Fluent and accurate speaking, one of the major parts of mastering a language, will greatly improve communication efficiency\cite{ex0}. The huge demand for spoken language learning has promoted the development of computer-assisted pronunciation training(CAPT) technology~\cite{chen2016computer,fouz2015trends}. Automatic speech assessment, as one of the key technologies of CAPT, can evaluate learners' speech across various dimensions such as accuracy, fluency, and completeness. Spoken language learning is typically divided into two scenarios: the open scenario and the follow-up scenario. In the follow-up scenarios, learners are required to follow and speak out a given prompt text, while the open scenarios allow learners to express opinions freely based on a posed question. In this paper, we focus on the sentence-level pronunciation scoring of accuracy and fluency in the follow-up scenario.

  
Most previous studies of sentence-level spoken speech evaluation fall into two categories, namely align-based~\cite{multipa,gopt,3m,our1,our2,wong22_interspeech,hipama} and align-free~\cite{liuwei,ssl1,liang23_interspeech}. Align-based systems require a DNN-HMM based Automatic speech recognition (ASR) model, which is used to perform forced alignment between the learners' speech and the corresponding prompt text. The alignment will get the phoneme sequence, the start and end times of each phoneme within the speech segment, and acoustic features extracted from the ASR system (such as hidden layer features or GOP features~\cite{gop} calculated by posterior probability). After obtaining the above features, a simple scoring network is built (usually Transformer~\cite{gopt,3m,hipama} or BLSTM~\cite{our1,our2,wong22_interspeech}) to map these features to the pronunciation scores. On the other hand, an align-free system is built with an end-to-end network, which directly maps the frame-level acoustic features and text-related features to scores. It significantly simplifies the feature extraction process and eliminates the impact of forced alignment on scoring. Past studies on align-free systems~\cite{liuwei,ssl1} have predominantly emphasized evaluating speech fluency~\cite{liuwei,ssl1} and prosody~\cite{ssl1}. In [13], the focus is on accuracy at both word and phoneme levels. Align-free system may have significant potential for sentence accuracy scoring of spoken language.

Recently, the emergence of large language models (LLMs) \cite{gpt4,llama,qwen} has demonstrated powerful language understanding and content generation capabilities, introducing some abilities that are not present in smaller-scale language models~\cite{llm_survey}, such as instruction following and in-context learning. Some previous research~\cite{text_score1,text_score2,text_score3,text_score4} has used LLMs to implement text-related scoring tasks. For example, the study~\cite{text_score3} asks the LLM first to analyze and interpret the provided materials, and then determine the final score. This method not only generates score results but also provides detailed comments on the student’s answers. In addition, the potential of LLM is not limited to text-related tasks. They can handle and understand non-text information such as audio and images well, thereby bridging the gap between different modalities. In the field of speech, the multi-modal models based on LLM usually consist of three main components: a speech encoder, a modality adapter, and a language model. Some representative multi-modal models~\cite{salmonn,wavllm,salm,qwen_audio}, such as SALMONN~\cite{salmonn}, Qwen-Audio~\cite{qwen_audio}, combine speech features and text embedding and feed them into a decoder-only language model, demonstrating powerful capabilities in speech understanding and recognition tasks.

Inspired by the above studies on multi-modal models and the positive effects of LLMs in text-related scoring tasks, we explore the use of multi-modal models for spoken pronunciation assessment tasks in this paper. The model training process is divided into two stages. The first stage is to train the model in the speech recognition task, and then the model will be fine-tuned based on the task-specific scoring data in the second stage. When assessing a learner's speech, the speech encoder was used to map the speech into contextual features. These features contain the prosody and content information in spoken language. Then the adapter layer maps this feature to the same space as the text embedding to achieve the modality adaptation. Finally, the prompt text and the assessment task-specific prefix are embedded and concatenated with the speech features generated by the adapter layer to allow the LLM to predict the accuracy and fluency scores. The contributions of this paper are summarized as follows:

\begin{figure*}[t!]
\centering
\includegraphics[width=0.6\textwidth]{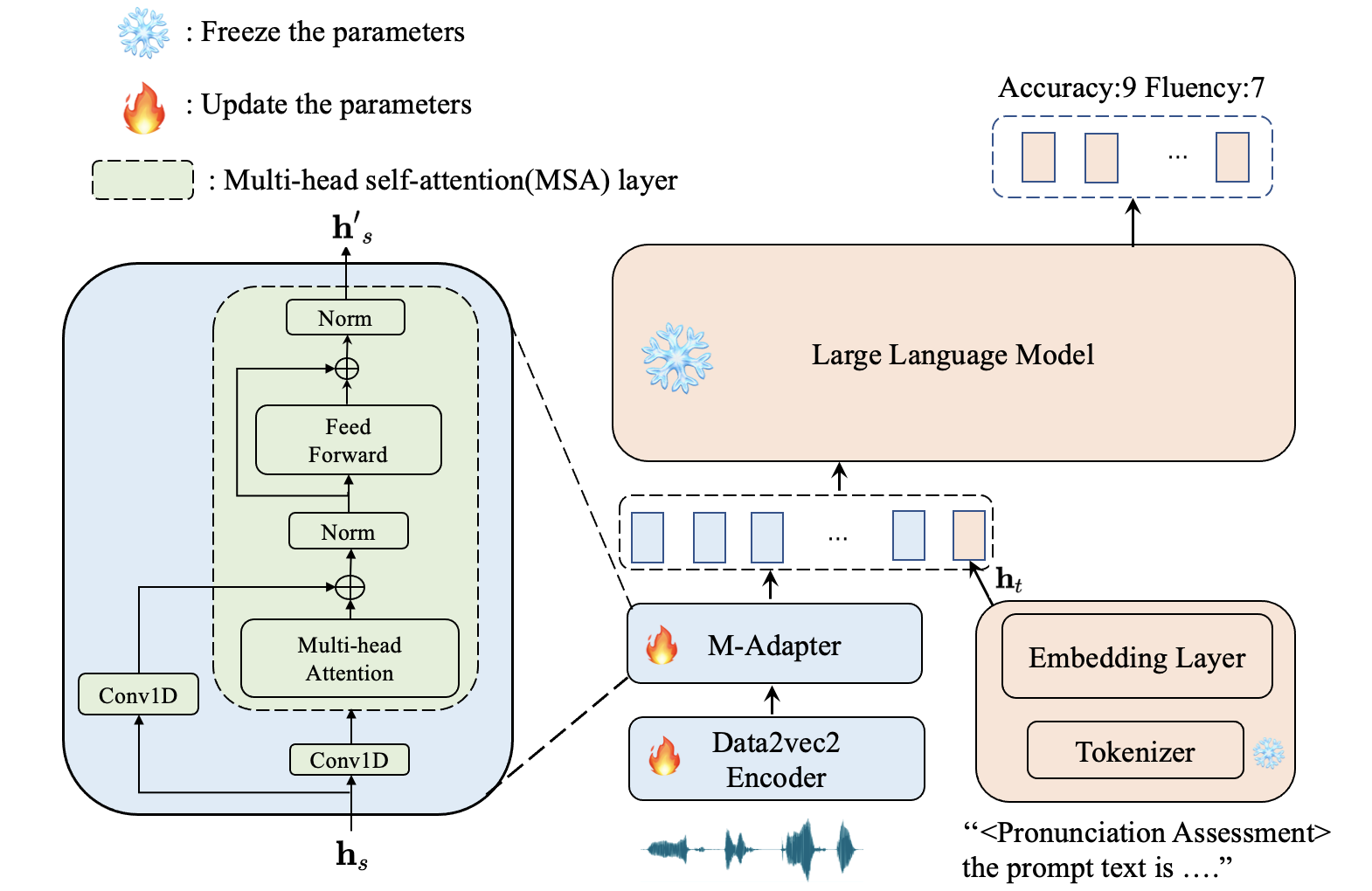}
\caption{The overall pipeline for L2 learner's pronunciation assessment in the proposed system.}
\label{fig: model}
\end{figure*}

\begin{itemize}
    \item To our knowledge, this paper is the first to use a multi-modality model based on the LLM for pronunciation assessment. 
    \item The proposed multi-modal method belongs to the align-free system category and has achieved competitive performance compared to traditional align-based and align-free systems, which also verifies the effectiveness of this method.
    \item We contributed to a multi-modal system that achieved state-of-the-art (SOTA) results in ASR when we finished the training in the first stage.
\end{itemize}

\section{Proposed Method}
This section describes the proposed multi-modal pronunciation assessment method. First, we outline the three components of our system: the pre-trained acoustic encoder from data2vec2, the m-adapter as a modality adapter, and the LLM-based pronunciation assessment module. Then, we introduce the two-stage training process. The overall pipeline for pronunciation assessment is illustrated in Figure~\ref{fig: model}. 

\subsection{Speech Encoder}
Data2vec2~\cite{data2vec2} was employed as the audio encoder because of its versatility, efficiency, and outstanding performance in speech tasks. It comprises several convolutional blocks and multi-layer transformers. The convolutions in each block have 512 channels with strides of (5, 2, 2, 2, 2, 2, 2) and kernel sizes of (10, 3, 3, 3, 3, 2, 2). It transforms 16 kHz raw audio into a sequence of embeddings with a stride of 20 ms and a receptive field of 25 ms.

For each sample, given the raw audio $\mathbf{x_s}$, the speech features $\mathbf{h_s}$ was firstly extracted from the last hidden layer through the data2vec2 based speech encoder, which can be written as:

\begin{equation}
\label{eq:speech_encoder}
\mathbf{h_s} =  {SpeechEncoder}(\mathbf{x_s}),
\end{equation}

\subsection{Modality adapter}
The m-adapter~\cite{m-adapter} was firstly introduced for speech translation tasks. We employed it as a modality adapter in our model to connect the speech and text modalities. It consists of two pooling convolutions and a multi-head self-attention (MSA) layer. The pooling operations are used to modify the speech sequence length, while the MSA captures global information.

This work follows the paper~\cite{seamless} by replacing the three independent pooling modules for Q, K, and V with a shared pooling module to streamline the architecture and improve the efficiency of the model. In general, given the speech features $\mathbf{h_s}$ extracted from the speech encoder, a convolutional pooling operation is first used to adapt the speech length. These adapted features are then fed into the MSA layers to obtain the modality adapter features $\mathbf{{h'}_s}$, expressed as follow:

\begin{equation}
\label{eq:adapter}
\mathbf{{h'}_s} =  {Adapter}(\mathbf{h_s}),
\end{equation}

The intuition of this module is that the number of features in the time domain of the speech signal is much larger than the length of the text. A scaling method needs to be applied to the speech features to balance the amount of information between the two modalities.

\begin{table*}[t!]
\vspace{1pt}
  \centering
  \caption{The WER results on Librispeech dataset comparing with other multi-modal systems. We briefly outlined the differences in these systems regarding speech encoders and modality adapters. Extra data indicates that these systems used additional training data besides Librespeech.}
   \label{table:asr_main}
\begin{tabular}{llcccc}
\toprule
      \multicolumn{1}{l}{\textbf{Model}}       & \textbf{Speech encoder} & \textbf{Modality adapter}    & \textbf{Extra data}  & \textbf{Test-clean} & \textbf{Test-other}   \\ \midrule
      SALMONN \cite{salmonn}           &    Whisper~\cite{whisper}, Beats   &  Q-former     &   \cmark             & 2.4                 & 5.4                        \\
      WavLLM~\cite{wavllm}     &  Whisper, WavLM          &  Linear &  \cmark         & 2              &  4.8                         \\  
      Qwen-Audio \cite{qwen_audio}  &  Whisper    & Linear &    \cmark   &   2.04     &  4.19                        \\ 
     SALM \cite{salm}  &     Hubert~\cite{hubert}   &  Linear &  \xmark       &   1.94        & 3.81                       \\ 
     Proposed   &     Data2vec2   & M-adapter & \xmark       &   \textbf{1.73}          &  \textbf{3.7}                       \\ 

\bottomrule

\end{tabular}
\end{table*}

\subsection{Pronunciation assessment with LLM}
Decoder-only LLM serves as a foundational component in the proposed method. We expect it to understand the learner's speech, analyse the difference between speech and the canonical text, and then make decisions regarding accuracy and fluency scores accordingly. 

For the text input of the LLM, we define the task-specific prefix as ``$<Pronunciation~Assessment>$". Additionally, in follow-up scenarios, The prompt text, as an important reference cue, can be considered as the text input of the LLM. Therefore, the final text input of LLM is defined as ``$<Pronunciation~Assessment> the~prompt~text~is~...$", and then the text input will be transformed into embedding features denoted as $\mathbf{{h}_t}$. Finally, we concatenate the embedding features with adapter-modified speech features $\mathbf{{h'}_s}$ and then fed the concatenated features into the LLM to obtain the final prediction text $\mathbf{y}$ through a token decoding operation as equation\eqref{eq:llm}, where the [;] denotes the concatenation of two vectors.

\begin{equation}
\label{eq:llm}
\mathbf{y} =  {LLM}([\mathbf{{h'}_s};\mathbf{h_t}]),
\end{equation}

\subsection{Two stages training  strategy}
The proposed method adopts a two-stage training strategy. Previous studies have shown that the performance of traditional align-based methods can be affected by the ASR system, as accuracy scoring heavily relies on speech content recognition. Therefore, in the first stage, we train the multi-modal model on the ASR task to make the connection between the two modalities more robust, which may benefit the pronunciation assessment task. We took the ASR task-specific prefix ``$<transcript>$" as the text input and then utilized limited ASR data to complete the ASR task training.

In the second stage, we fine-tuned the model trained in the first stage using a limited amount of scoring data based on the ``$<pronunciation~assessment>$" assessment task-specific prefix and prompt text. For the ground truth, we structured it as ``$accuracy:9~fluency:7$", aiming for LLM to output scores in such a format to achieve multi-task scoring for both accuracy and fluency.

\section{Experimental setup}

\subsection{Datasets}
Two public datasets were used in this study. Librispeech~\cite{librispeech}, a commonly used ASR corpus, was utilized to train the multi-modal ASR in the first stage. It consists of about 1,000 hours of 16kHz read English speech derived from read audiobooks, with the dev-clean and dev-other subsets used as the validation set, and the test-clean and test-other subsets used as the test sets. To evaluate fluency and accuracy scoring in the second stage, we conducted experiments on Speechocean762~\cite{speechocean}, an open-source speech assessment corpus consisting of 5,000 English utterances collected from 250 Chinese speakers. The whole scoring dataset is divided into 2,500 sentences for training and 2, 500 for testing. The score distribution of the test set is shown in Figure 2. The x-axis represents the human-labeled scores, and the y-axis represents the number of occurrences for each corresponding score. 

\begin{figure}[t!]
\centering
\includegraphics[width=0.35\textwidth]{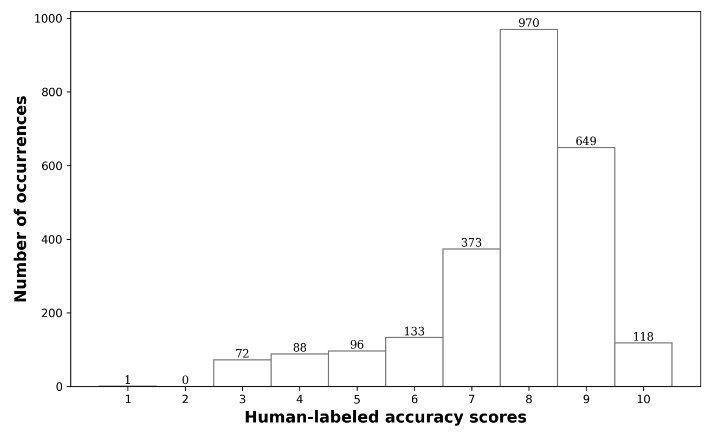}
\includegraphics[width=0.35\textwidth]{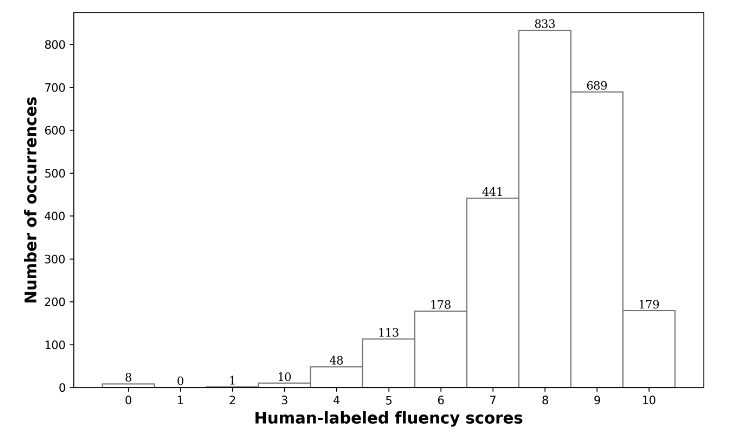}
\caption{The distribution of the human-labeled scores}
\label{fig: model}
\end{figure}


\subsection{Setup}

The weight of the data2vec2-based speech encoder is initialized from data2vec2-large~\footnote{\scriptsize{https://dl.fbaipublicfiles.com/fairseq/data2vec2/large\_vox\_960h.pt}} and the Qwen-based LLM~\cite{qwen} is initialized using pre-trained weights derived from Qwen-7B~\footnote{\scriptsize{https://huggingface.co/Qwen/Qwen-7B}}. For the M-adapter, we chose one MSA layer with 16 heads. The kernel size and stride of the convolution were both set to 2, aiming to halve the length of speech features extracted from the speech encoder.  

We used 8 A800s for the experiments. Each stage was trained for a maximum of two epochs on the corresponding task data. The ASR and pronunciation scoring training took about 9 and 0.2 hours, respectively. During both stages, the parameters of LLM were frozen, and the parameters of the speech encoder and modality adapter were updated.

\section{Results and Discussion}
This section first presents the ASR result of the proposed system from the first training stage. Then, we demonstrate the pronunciation assessment results after fine-tuning the model based on the ASR training. Additionally, we present ablation experiments to explore the impacts of ASR training and prompt text. The scoring system performance was evaluated using the Pearson correlation coefficient (PCC) between the predicted scores and the human-labeled scores.

\subsection{Performance of ASR}
ASR task was often used to evaluate the performance of multi-modal systems\cite{salmonn,wavllm,qwen_audio,salm}. In this subsection, we selected several representative multi-modal systems with strong ASR performance as our baselines, including SALMONN~\cite{salmonn}, WavLLM~\cite{wavllm}, Qwen-Audio~\cite{qwen_audio}, and SALM~\cite{salm}. 

\begin{table*}[t!]
\vspace{1pt}
  \centering
  \caption{The PCC results on Speechocean762 dataset comparing with other baseline systems.}
   \label{table:main_result}
\begin{tabular}{llcccc}
\toprule
      \multicolumn{1}{l}{\textbf{Type}}   & \textbf{Methods}    & \textbf{Fluency} & \textbf{Accuracy}     \\ \midrule
       \multirow{3}{*}{Align-based}  &   GOPT~\cite{gopt}           &    0.753   &  0.714                        \\
                                    &    Multi-task~\cite{wong22_interspeech}      &  0.73          &  0.694                 \\  
                                    &   MultiPA~\cite{multipa}  &  0.772    & 0.705           \\   
                                    &   HiPAMA \cite{hipama}  &  0.762    & 0.730           \\   \midrule
       \multirow{3}{*}{Align-free}   & SSL~\cite{ssl1}  &     0.78   &  -          \\ 
                                     & SSL+IDX~\cite{liuwei}  &     0.795   &  -          \\ 
                                    &  \textbf{Proposed}   &     \textbf{0.777}   & \textbf{0.713}                 \\ 

\bottomrule

\end{tabular}
\end{table*}

\begin{table}[ht]
  \centering
  \caption{The WER results of speech encoder models in the proposed method.}
   \label{table:asr1}
\begin{tabular}{clcccc}
\toprule
      \multicolumn{1}{l}{\textbf{Method}}   & \textbf{Speech encoder}  & \textbf{Test-clean} & \textbf{Test-other}   \\  
   \midrule
   \multirow{3}{*}{Proposed}  &  Whisper        & 2.28              & 6.34                 \\
                              &  Hubert       & 1.83             &  \textbf{3.68}                \\ 
                              &  \textbf{Data2vec2}      & \textbf{1.73}              & 3.7                 \\ 
                              
\bottomrule
\end{tabular}
\end{table}

The comparison results between the proposed system and these baseline systems are shown in Table~\ref{table:asr_main}. For the result of the proposed method, the predict and ground-truth texts were normalized using Whisper's text normalization method~\footnote{\scriptsize{https://github.com/openai/whisper/tree/main/whisper/normalizers}} before computing WER. The results suggest that the proposed method outperforms all the baselines in terms of WER in the Librispeech test sets. This observation encourages us greatly for our upcoming pronunciation assessment tasks.

In addition, we briefly compared several different speech encoders based on the proposed method, with results shown in Table~\ref{table:asr1}. The findings reveal that the data2vec2 model as the optimal choice for speech encoding performs significantly better compared to the performance of Hubert~\footnote{\scriptsize{https://huggingface.co/facebook/hubert-large-ls960-ft}} and Whisper~\footnote{\scriptsize{https://huggingface.co/openai/whisper-large-v2}}.


\subsection{Results of pronunciation assessment}
In this subsection, we present the performance of the proposed multi-modal model in pronunciation scoring, which is an align-free end-to-end approach. For a fair comparison, we selected two types of baseline systems: one is align-free end-to-end systems based on SSL features and the other one is align-based systems that rely on traditional DNN-HMM based ASR model trained with Librispeech dataset. We did not consider the align-based system using SSL features~\cite{3m} as our baseline, since such an approach would make the entire align-based pipeline more complicated and less feasible for practical implementation.

We first conducted performance comparisons between the proposed multi-modal pronunciation scoring system and the align-based baseline systems presented in Table~\ref{table:main_result}. The leading results validate the effectiveness of our proposed method.
Moreover, compared to the complex pipelines of align-based systems, our proposed method offers a simpler way to evaluate learners' pronunciation. 

Then, we evaluate the performance comparisons with the align-free methods. The proposed method demonstrates competitive results with baseline systems in fluency scoring. Additionally, it fills the gap in accuracy scoring for align-free approaches on the Speechocean762 dataset.



\subsection{Ablation studies}
In this subsection, we conducted experiments to determine the relative importance of ASR training in the first stage and prompt text in the proposed system for pronunciation scoring. The results are presented in Table~\ref{table:ablation}.

Firstly, we can clearly see that these two factors slightly improve fluency scoring performance in system (a). The two factors focus more on the content of the speech. As verified in \cite{our1,our2}, prosodic information is more important than speech content information for fluency scoring. Moreover, compared to SSL system~\cite{ssl1}, SSL+IDX~\cite{liuwei}showed a slight performance difference due to the additional information provided by phoneme IDs, which represent speech content. This further confirms that the content does not have a significant impact on fluency scoring.

Compared to speech fluency, accuracy is more significantly affected by these factors. The results of systems (b) and (c) show that the impact of introducing the prompt text is greater than that of ASR training. Additionally, system (d) found that not using any strategies resulted in poor accuracy scoring performance.

\begin{table}[ht]
\vspace{1pt}
  \centering
  \caption{The PCC results of ablation studies for the proposed method.}
   \label{table:ablation}
\begin{tabular}{llcccc}
\toprule
            &      \textbf{Factors}    & \textbf{Fluency} & \textbf{Accuracy}     \\ \midrule
       (a) & ASR training + Prompt text   &      \textbf{0.777}          &  \textbf{0.713}                 \\  
      (b) &         ASR training only      &   0.763    & 0.698           \\   
      (c) &          Prompt text only     &   0.768    & 0.665          \\  
      (d) &                      -      &    0.759    & 0.654          \\  

\bottomrule

\end{tabular}
\end{table}

\section{Conclusions}
This paper introduces the multi-modal approach to develop an end-to-end pronunciation assessment system for scoring fluency and accuracy of learners' L2 speech, such that it does not require any word and phone-level alignment information as input. The experiment results show that our approach achieves a competitive performance compared with previous methods, including those systems that utilize alignment information. This work can be used as the baseline system of end-to-end multi-modal approach and align-free related approach for pronunciation assessment tasks. Additionally, our proposed method achieved state-of-the-art (SOTA) performance in the ASR task, demonstrating the superior effectiveness of our multi-modal system. 

In the future, we will investigate the chain-of-thought (COT)~\cite{cot} ability of the proposed system to provide step-by-step explanations for the model's scoring decisions.

\bibliographystyle{IEEEbib}

\small
\bibliography{strings,main}

\end{document}